\documentclass[10pt,twocolumn,letterpaper]{article}

\usepackage[pagenumbers]{wacv} % To force page numbers, e.g. for an arXiv version

\usepackage{graphicx}
\usepackage{amsmath}
\usepackage{amssymb}
\usepackage{booktabs}
\usepackage{multirow}
\usepackage{xcolor}
\usepackage{soul}
\usepackage{booktabs}
\usepackage{pifont}

\usepackage[pagebackref,breaklinks,colorlinks]{hyperref}

\usepackage[capitalize]{cleveref}
\crefname{section}{Sec.}{Secs.}
\Crefname{section}{Section}{Sections}
\Crefname{table}{Table}{Tables}
\crefname{table}{Tab.}{Tabs.}

\def\wacvPaperID{1881} % *** Enter the WACV Paper ID here
\def\confName{WACV}
\def\confYear{2025}

\begin{document}

\title{VLTP: Vision-Language Guided Token Pruning for Task-Oriented Segmentation}

\author{Hanning Chen$^{1}$\thanks{This work was done during an Cisco internship.} \quad Yang Ni$^{1}$ \quad Wenjun Huang$^{1}$ \quad Yezi Liu$^{1}$ \quad SungHeon Jeong$^{1}$ \\ 
\quad Fei Wen$^{2}$ \quad Nathaniel Bastian$^{3}$ \quad Hugo Latapie$^{4}$\thanks{Internship manager} \quad Mohsen Imani$^{1}$ 
\vspace{0.3em} 
\\
{\normalsize $^1$ University of California, Irvine, CA, USA} \\
{\normalsize $^2$ Texas A\&M University, College Station, TX, USA} \\
{\normalsize $^3$ United States Military Academy, West Point,  USA} \\
{\normalsize $^4$ Cisco, San Jose, CA, USA} \\
{\tt\small \{hanningc, m.imani\}@uci.edu}
% {\normalsize $^3$Department of Computing, The Hong Kong Polytechnic University}
}

\maketitle

%%%%%%%%% ABSTRACT
\begin{abstract}
Vision Transformers (ViTs) have emerged as the backbone of many segmentation models, consistently achieving state-of-the-art (SOTA) performance. However, their success comes at a significant computational cost. Image token pruning is one of the most effective strategies to address this complexity. 
However, previous approaches fall short when applied to more complex task-oriented segmentation (TOS), where the class of each image patch is not predefined but dependent on the specific input task.
% Previous approaches have either predicted the semantic class of simpler image patches early or merged patches with the same semantic class. 
% However, these methods fall short when applied to more complex task-oriented segmentation (TOS), where the class of each image patch is not predefined but dependent on the specific input task.
% Additionally, recent advancements in visual reasoning have explored integrating multi-modal large language models (MLLM) to generate prompts or provide guidance, thereby enhancing the performance of vision models in segmentation tasks. 
% Therefore, there is a pressing need for a novel token pruning mechanism that can accelerate ViT-based segmentation models, particularly for TOS guided by MLLM. 
This work introduces the \underline{\textbf{V}}ision \underline{\textbf{L}}anguage Guided \underline{\textbf{T}}oken \underline{\textbf{P}}runing (VLTP), a novel token pruning mechanism that can accelerate ViT-based segmentation models, particularly for TOS guided by multi-modal large language model (MLLM).
% The core idea is that not all image tokens need to be processed by all layers of the ViT—only those related to reasoning tasks. 
We argue that ViT does not need to process every image token through all of its layers—--only the tokens related to reasoning tasks are necessary.
We design a new pruning decoder to take both image tokens and vision-language guidance as input to predict the relevance of each image token to the task. Only image tokens with high relevance are passed to deeper layers of the ViT. Experiments show that the VLTP framework reduces the computational costs of ViT by approximately 25\% without performance degradation and by around 40\% with only a 1\% performance drop. The code associated with this study can be found at \href{https://github.com/HanningChen/VLTP/tree/main}{this URL}.      
\vspace{-1.5em}
\end{abstract}

%%%%%%%%% BODY TEXT
\section{Introduction}
ViTs~\cite{dosovitskiy2020image} and multi-head self-attention (MHSA)~\cite{vaswani2017attention} have significantly advanced the development of computer vision, where ViTs have been widely used as backbone models for various tasks~\cite{khan2022transformers}, including image classification~\cite{caron2021emerging} and segmentation~\cite{kirillov2023segment}.
Compared to previous convolutional neural networks (CNN)~\cite{he2016deep, huang2024recoverable, huang2024intelligent, zhang2024advancing}, ViT-based models excel in global reasoning through pairwise token attention. However, the success of ViT models comes with a significant computational overhead, which limits their deployment in resource-constrained environments, such as edge computing. This challenge is especially pronounced in segmentation tasks. 
% , where high-resolution images generate numerous input tokens.
Since segmentation requires detailed pixel-wise predictions, they often use high-resolution images, further increasing the number of image patches and leading to a quadratic increase in computation. 

\begin{figure}[t]
  \centering \includegraphics[width=0.85\columnwidth]{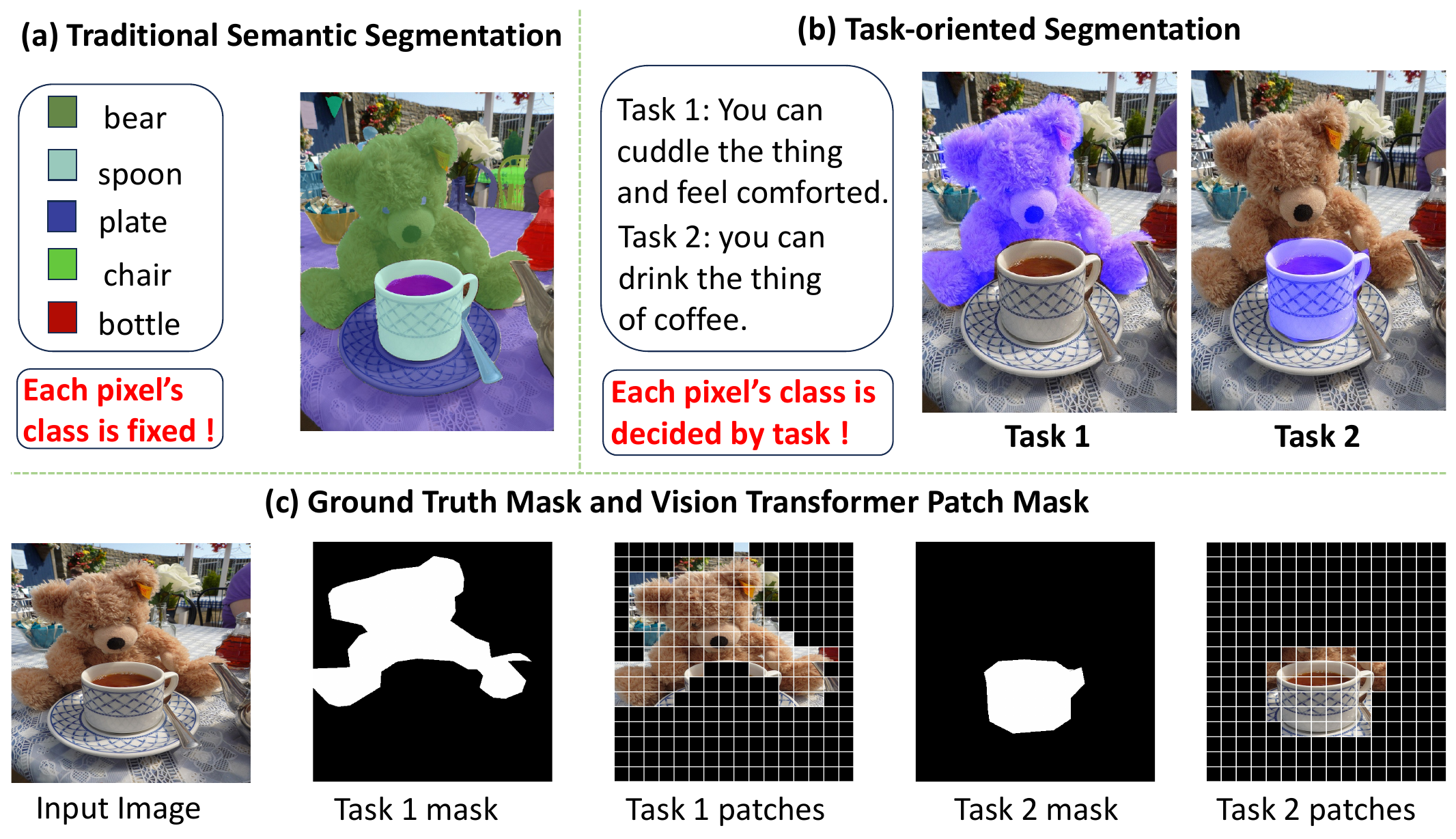}
   \caption{(a) Semantic segmentation example. (b) TOS example. (c) For the same image, the segmentation mask and corresponding image patches change when the input task changes.}
   \label{fig:motivation}
   \vspace{-1.5em}
\end{figure}

To improve the efficiency of ViT models, several techniques have been proposed~\cite{chen2024comprehensive}.
Among the most notable methods is image token pruning~\cite{tang2023dynamic,lu2023content}, which effectively accelerates model inference. These approaches aim to reduce the number of image tokens involved in ViT layers such as the MHSA computation, thus significantly decreasing the computational load due to the inherent quadratic complexity \cite{vaswani2017attention}. %For example, the DToP method~\cite{tang2023dynamic} intends to predict the semantic class of simpler image tokens in the early ViT layers. 
For example, the DToP method~\cite{tang2023dynamic} aims to predict the semantic class of image tokens in the early layers of ViT, and those with high confidence can be omitted from computations in deeper ViT layers. Alternatively, CTS~\cite{lu2023content} aims to combine image tokens that share the same semantic class, thereby reducing the total token count. As illustrated in~\cref{fig:motivation}(a), these techniques work well for traditional segmentation tasks. However, they are less effective for more complex vision language tasks such as task-oriented segmentation (\textbf{TOS})~\cite{sawatzky2019object,qu2024rio}, where the class of each pixel varies depending on the specified task.
As seen in~\cref{fig:motivation}(b), TOS involves finding an item's mask based on the query task. As the task changes, so does the ground truth mask, meaning that each pixel's class also changes, as shown in~\cref{fig:motivation}(c). Directly applying previous methods~\cite{tang2023dynamic,lu2023content} would lead to a significant loss of visual information and thus low-quality masks during the reasoning stage, because these methods focus on the relevance of image patches to static semantic class, without considering a specific task. A further drawback of earlier approaches is their incompatibility with the latest trending MLLMs~\cite{liu2024visual} since their pruning strategies did not account for external reasoning guidance. Recently, numerous frameworks~\cite{lai2024lisa,qian2024affordancellm,ma2024groma,zhang2024groundhog} have been proposed to guide vision models in performing vision language tasks using MLLM, such as Referring Expression Segmentation (RES)~\cite{yu2016modeling}. The strong reasoning capability of MLLM makes it promising for solving TOS tasks. Therefore, it is necessary to develop a new token pruning mechanism to accelerate ViT-based segmentation models, especially for TOS tasks guided by MLLM.

In this research, we present VLTP (\underline{\textbf{V}}ision-\underline{\textbf{L}}anguage Guided \underline{\textbf{T}}oken \underline{\textbf{P}}runing), a new ViT token pruning technique for TOS. The core concept of VLTP is that \textbf{only tokens relevant to reasoning tasks should be processed by all ViT layers}. We design a novel prune decoder and insert it at multiple selected layers of ViT to provide flexible multi-stage pruning. The prune decoder predicts the relevance of image tokens to the reasoning tasks, leveraging the guidance from MLLM. Based on the relevance predictions from the prune decoder, VLTP retains only the image tokens with high relation scores and forwards them to the following layers of ViT. By removing unrelated tokens that participated in ViT computation, we effectively accelerate the ViT model while preserving its high reasoning efficacy. To mitigate mispredictions by the prune decoder (e.g., due to pruning in the early stages of ViT), image tokens dropped are reactivated in the next pruning stage for re-evaluation of token relevance as well as in the mask generation. The contributions of the work are summarized as follows:

\begin{itemize}
\item We propose a token pruning strategy aimed at TOS with ViT-based models. By leveraging guidance from MLLM, our approach removes image tokens with low task relevance from later computation, thereby forcing the model to concentrate on important image tokens pertinent to the task. %To address mispredictions, we introduce a reactivation mechanism to reassess the connections of all image tokens to the reasoning tasks.
\item To the best of our knowledge, this is the first ViT token pruning method that takes MLLM guidance into consideration. Unlike previous works that focus on traditional vision-only tasks (such as image classification and segmentation), this work is also the first to investigate ViT pruning in vision-language reasoning tasks.
\item Experiments on visual reasoning tasks show that VLTP reduces the computational costs of ViT by approximately 25\% without any performance degradation and by around 40\% with only a 1\% performance drop. 
Guided by MLLM, our VLTP-integrated segmentation model outperforms the SOTA methods, achieving a +2.5\% improvement in mIoU.

% \textcolor{red}{\st{Under MLLM guidance, a VLTP integrated segmentation model outperforms the SOTA methods, achieving a +2.5\% mIoU on common tasks and +2.3\% on uncommon tasks when conducting challenging visual reasoning tasks.}
% }
\end{itemize}

\section{Related Works}
\subsection{Image Segmentation}
Image segmentation is the process of dividing an image into distinct regions or segments, each representing different objects or areas of interest. Traditional segmentation tasks include instance segmentation~\cite{bolya2019yolact,wang2023cut,he2023fastinst,wang2021end} and semantic segmentation~\cite{strudel2021segmenter,cheng2021per,xie2021segformer,caesar2018coco,mottaghi2014role}. Instance segmentation identifies and separates individual objects in an image, assigning a unique label to each object instance. Semantic segmentation, on the other hand, classifies every pixel, partitioning the image into meaningful regions based on visual characteristics. These traditional segmentation tasks are typically image-based, where pixels are categorized into predefined classes, as shown in~~\cref{fig:motivation}(a). 
More recently, segmentation tasks that incorporate both image and text inputs have emerged, such as RES~\cite{lu202012} and affordance detection~\cite{cuttano2024does}. These tasks involve identifying object masks using a specified text query. Building on these approaches, more complex task-driven~\cite{sawatzky2019object} and intention-driven~\cite{qu2024rio} segmentation tasks have been introduced, focusing on locating appropriate items to accomplish a given task (or intention, which is a more abstract task). An example of Task-Oriented Segmentation (TOS) is shown in~\cref{fig:motivation}(b). These novel TOS tasks require models to perform not only visual recognition and scene comprehension but also reasoning.

\subsection{Vision Transformer for Segmentation}
ViT has become a popular backbone network for segmentation tasks~\cite{zheng2021rethinking,xie2021segformer,strudel2021segmenter}. Compared to CNN-based segmentation models~\cite{jiang2021semantic}, the global attention mechanism~\cite{vaswani2017attention} enables ViTs to better understand complex scenes and achieve higher segmentation accuracy, in scenarios with intricate object relationships. The ViT-backbone also facilitates the integration of text embedding information, allowing the model to leverage additional semantic context for more accurate and complex segmentation task~\cite{liu2024visual,zhang2022glipv2,kirillov2023segment}. However, the computational cost of ViT limits its deployment, particularly in high-resolution image segmentation tasks. The need for deeper networks and numerous image tokens makes model compression essential.  
\vspace{-0.3em}
\subsection{Token Pruning}
Token pruning has been shown to be an effective mechanism for accelerating ViT~\cite{kong2022spvit,liu2024revisiting,mahmud2024papr,rao2021dynamicvit}. The key idea behind token pruning is the sparsification of image tokens. Since ViT’s attention computation latency and memory requirements are quadratic to the sequence length of the image tokens~\cite{katharopoulos2020transformers}, reducing the number of tokens involved in computation has proven to be an effective method for accelerating ViT. Existing ViT token pruning techniques have been successful in image classification~\cite{yin2022vit,pan2021ia} and traditional image segmentation tasks~\cite{tang2023dynamic,liu2024revisiting,lu2023content}, but they are difficult to apply in TOS tasks. %Previous ViT token pruning methods rely on estimating the semantic class of tokens and either dropping easily predictable tokens or merging tokens that share the same semantic class. 
As depicted in~\cref{fig:motivation}(c), which highlights the complexity in TOS tasks, the label assigned to a pixel varies depending on the task. That is, the importance of pixels is ambiguous without extra information. Therefore, a new pruning mechanism is needed. % Therefore, a new pruning mechanism is needed that considers both image visual information and task-based reasoning, enabling the acceleration of the ViT while ensuring the model generates high-quality masks.

\section{Task-oriented Segmentation}
 
\begin{figure}[t]
  \centering \includegraphics[width=0.9\columnwidth]{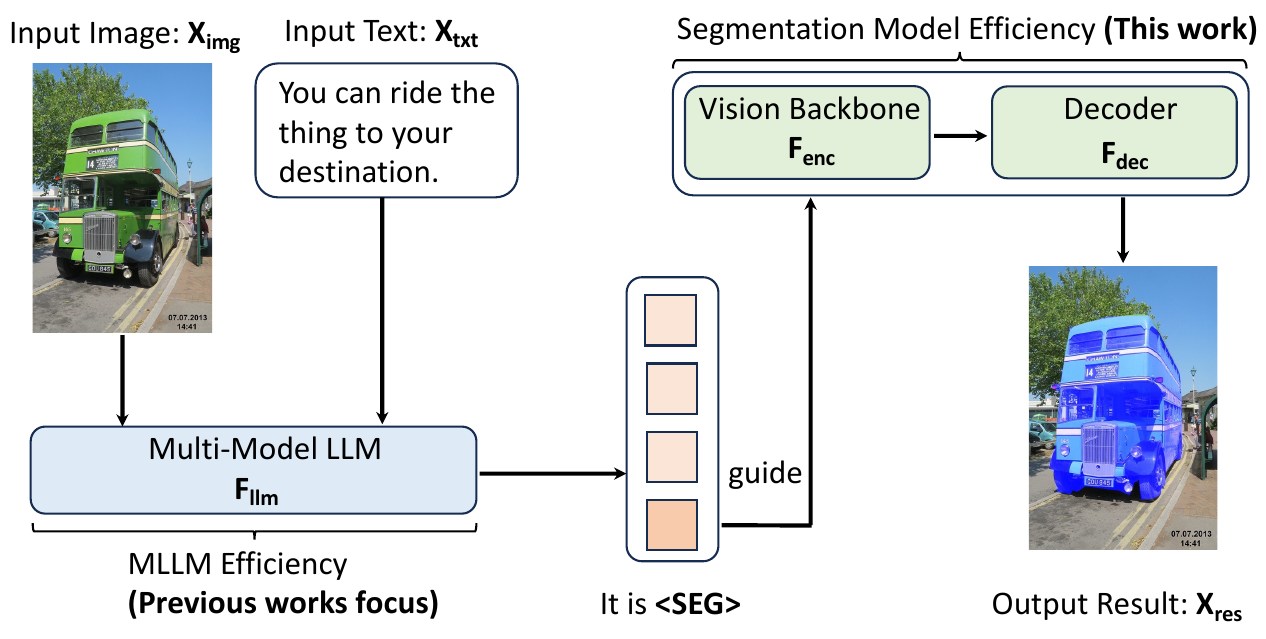}
   \caption{Multi-model LLM (MLLM) guide segmentation model for TOS.}
   \label{fig:MLLM_arch}
   \vspace{-1em}
\end{figure}

% \begin{figure}[t]
%   \centering \includegraphics[width=\columnwidth]{figures/compare_previous_work.pdf}
%    \caption{(a) Comparison with previous works. (b) MLLM guide segmentation model latency breakdown.}
%    \label{fig:table}
%    \vspace{-1.5em}
% \end{figure}

% \usepackage{multirow}
% \usepackage{graphicx}

\begin{table}[]
\centering
\resizebox{0.9\columnwidth}{!}{%
\begin{tabular}{ccc}
\toprule
\multicolumn{3}{c}{\textbf{(a) Comparison with Previous Works}}                                                   \\ \midrule
\multicolumn{1}{c|}{Previous works} &
  \multicolumn{2}{l}{\begin{tabular}[c]{@{}l@{}}1. MLLM   quantization, Pruning, and KG distillation.\\ 2. ViT pruning without LLM guidance.\end{tabular}} \\ \midrule
\multicolumn{1}{c|}{This work}                     & \multicolumn{2}{l}{ViT pruning under LLM guidance.} \\ \midrule \midrule
\multicolumn{3}{c}{\textbf{(b) Latency Breakdown (single image and single text)}}                                 \\ \midrule
\multicolumn{1}{c|}{\multirow{2}{*}{MLLM (LLaVA 7B)}} & \multicolumn{2}{c}{Segment Anything (SAM)}          \\ \cline{2-3} 
\multicolumn{1}{c|}{}                              & \multicolumn{1}{c|}{ViT-H}            & Mask Decoder       \\ \midrule
\multicolumn{1}{c|}{71.0 ms}                       & \multicolumn{1}{c|}{97.7 ms}        & 8.9 ms        \\ \midrule
\multicolumn{1}{c|}{40\%}                          & \multicolumn{1}{c|}{55\%}           & 5\%           \\ \bottomrule
\end{tabular}%
}
\caption{(a) Comparison with previous works. (b) MLLM guide segmentation model latency breakdown. We conudct the benchmark using an RTX A6000 GPU with the RIO dataset~\cite{qu2024rio}.}
\label{tab: motivate}
\end{table}

%\subsection{Two-stage Framework}
\noindent \textbf{Problem Definition:} The TOS task is formulated as follows: with an input image $X_{img}$ and a task $X_{txt}$, we need to find the segmentation mask $X_{res}$ for all suitable affordances. 
\begin{equation}
    X_{res} = \mathbf{F}(X_{img}, X_{txt})
\end{equation}
Here \textbf{F} refers to the model to solve the task. As shown in~\cref{fig:MLLM_arch}, to solve the task ``\textit{You can ride the
thing to your destination}'', we need to find the mask of the object ``\textit{bus}''. In this paper, we propose to leverage a nature paradigm to solve this task. We first query an MLLM, such as LLaVA~\cite{liu2024visual}, to perform reasoning over the image based on the task and generate a special token $\langle$\text{SEG}$\rangle$.
\begin{equation}
    \langle SEG \rangle = \mathbf{F}_{LLM}(X_{img}, X_{txt})
\end{equation}
Next, the special token is treated as a pointer to guide the segmentation model (such as Segment Anything (SAM)~\cite{kirillov2023segment}) to find the mask. 
\begin{equation}
    X_{res} = \mathbf{F}_{seg}(X_{img}, \langle SEG \rangle)
\end{equation}
, where \textbf{F}$_{seg}$ is the segmentation model. 

%\subsection{Comparison to Prior Works}
% In contrast to conventional one-stage vision-language models (such as MDETR~\cite{kamath2021mdetr} and SeqTR~\cite{zhu2022seqtr}), our two-step framework seamlessly integrates the robust reasoning skills of MLLMs with the exceptional segmentation quality of large vision foundation models. Though previous works~\cite{tang2023cotdet,chen2024taskclip}. 

\noindent \textbf{Comparison to Earlier TOS Solutions:} In addressing reasoning-intensive task-oriented object detection and segmentation tasks (e.g., COCO-Tasks~\cite{sawatzky2019object} and RIO~\cite{qu2024rio} datasets), two main types of solutions exist: single-stage and two-stage frameworks. Single-stage frameworks, such as MDETR~\cite{kamath2021mdetr} and Polyformer~\cite{liu2023polyformer}, directly combine image features and text features and perform end-to-end training of the entire system. The drawback of single-stage frameworks is their lack of logical reasoning capabilities. Consequently, recent works like CoTDet~\cite{tang2023cotdet} and TaskCLIP~\cite{chen2024taskclip} introduce two-stage frameworks that combine the reasoning abilities of LLMs with the object detection and segmentation capabilities of vision models. However, these frameworks require LLMs to explicitly generate task-dependent feature words and rationale. Sometimes the LLM provides only a vague indicator, such as a special token embedding, rather than a clear rationale. In contrast to these previous works, our two-step framework seamlessly integrates the robust reasoning abilities of MLLMs with the superior segmentation quality of vision foundation models.    

\textbf{Nevertheless, a key challenge remains: the computational expense of two-stage framework is extremely high.} Previous research~\cite{shang2024llava, ye2024voco, zhou2024tinyllava} has primarily focused on MLLM's compression and pruning, largely overlooking the segmentation model, as depicted in~\cref{tab: motivate}.(a). The segmentation model, as shown in~\cref{tab: motivate}(b), requires even more computational resources than MLLM. Following this, we will explore ways to enhance the efficiency of the segmentation model under the guidance of MLLM.

\section{Vision-Language Guided Patch Pruning}

\begin{figure*}[t]
  \centering \includegraphics[width=0.8\textwidth]{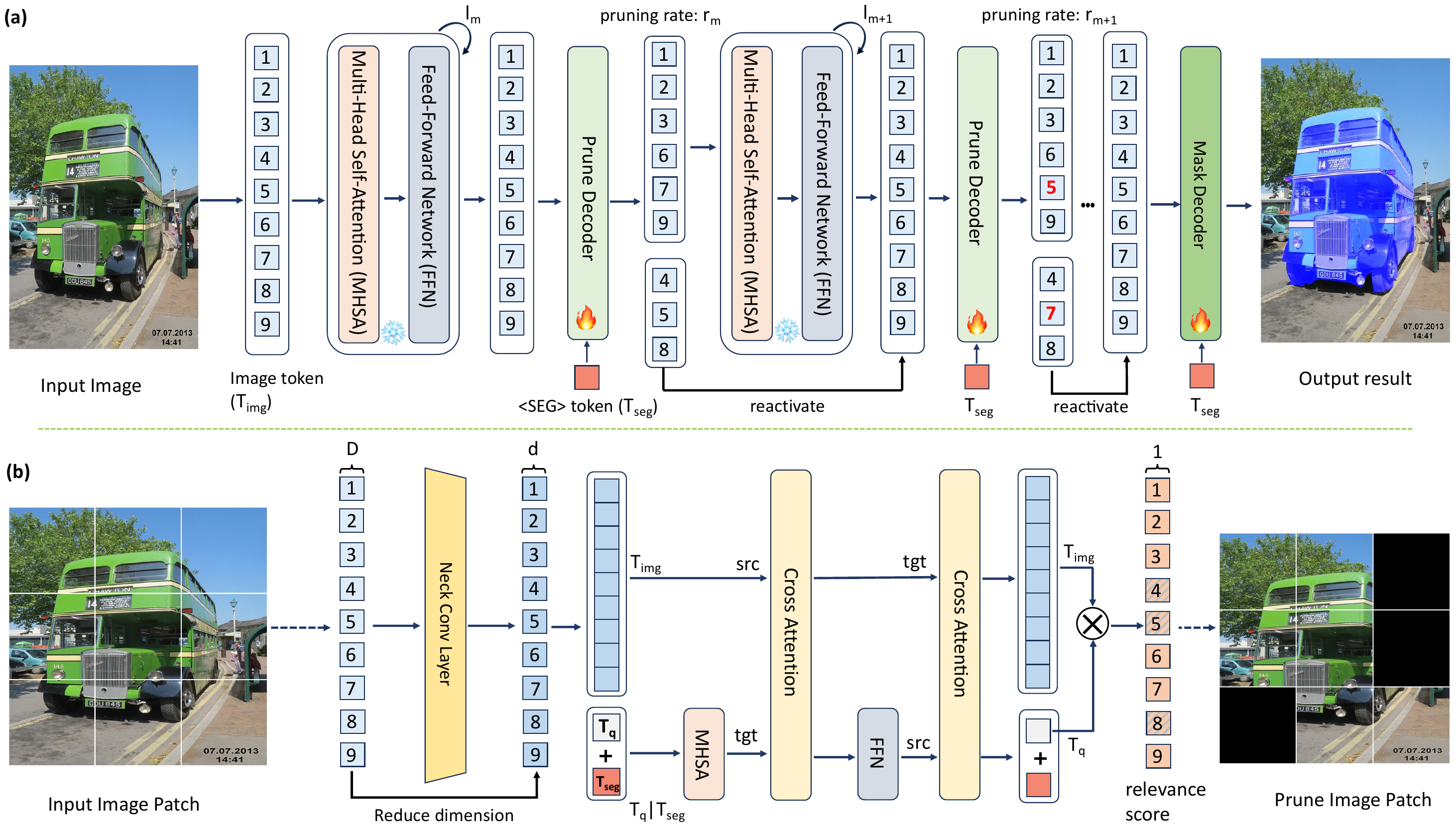}
   \caption{The illustration of  vision-language guided token pruning (\textbf{VLTP}) framework. (a) ViT architecture with prune decoder. (b) Prune decoder model architecture. In this illustration, we consider an image with only 9 patch tokens.}
   \label{fig:VLP_arch}
\end{figure*}

As illustrated in~\cref{tab: motivate}(b), the ViT's computation latency constitutes more than 90\% of the total latency of the entire segmentation model. Thus, enhancing the ViT's efficiency is essential for optimizing the segmentation model's performance. This work introduces a vision-language guided token pruning method, which accelerates ViT when handling TOS. We detail the mechanism in the following sections. 

\subsection{Preliminary: Segmentation Model}
Refer to~\cref{fig:MLLM_arch}, a segmentation model $\textbf{F}_{seg}$ typically consists of two components: a vision backbone $\textbf{F}_{enc}$ and a mask decoder $\textbf{F}_{dec}$. We illustrate this with SAM. The vision backbone in SAM is a ViT~\cite{dosovitskiy2020image,he2022masked}. The ViT extracts features from the input image and passes the feature map to the mask decoder. The mask decoder then predicts the target mask based on the feature map and prompt embeddings. Within our system, the prompt embedding is represented by the $\langle$\text{SEG}$\rangle$ token embedding from MLLM, which serves as an ambiguous indicator towards the intended affordance to solve the task. 
As depicted in~\cref{tab: motivate}(b), given the large latency of vision backbone, the segmentation model will benefit significantly from our proposed VLTP.    

A ViT splits an image ($X_{img}$ $\in$ $R^{3 \times H \times W}$) into different patches via patch embedding layer. Suppose the embedding dimension is $C$, the patch size is $P$, and the image token sequence length is $N$, we have:
\begin{equation}
    N = \frac{H \times W}{P^2}
\end{equation}
ViTs are position-agnostic, so we add positional encoding to represent the spatial information of the image token. The resulting image token sequence is denoted as $T_{img}^0 \in R^{N \times C}$. We then pass the image token sequence into multiple repeated layers. Every ViT layer contains an MHSA, a feed-forward network (FFN), a layer normalization (LN)~\cite{ba2016layer}, and a residual connection~\cite{he2016deep}. The layers are indexed by $l \in \{1, 2, ..., L\}$, and the output of each layer is marked as $T_{img}^l$.

\begin{equation} \label{eq:self-attention}
\begin{aligned}
    & T_{img}^{l'} = MHSA(LN(T_{img}^{l-1})) + T_{img}^{l-1} \\
    & T_{img}^{l} = FFN(LN(T_{img}^{l'})) + T_{img}^{l'}
\end{aligned}
\end{equation}

Given that the computation in~\cref{eq:self-attention} must be executed by $L$ times, the computational burden of ViT is exceptionally high, especially when generating high-quality masks. The ViT model is particularly computation-intensive in terms of layer depth and embedding dimension. For example, SAM's ViT-H model comprises 32 layers with an embedding dimension of 1280. Traditional methods to enhance the ViT's efficiency include quantization~\cite{li2023vit}, knowledge distillation~\cite{zhang2024efficientvit}, and model weight pruning~\cite{yang2023global}. Nonetheless, in this work, we present a novel orthogonal approach: \textbf{Not all image tokens need to traverse all ViT layers; we can eliminate the irrelevant tokens midway.}  

\subsection{Token Pruning Mechanism}
To keep the image tokens that are relevant to reasoning tasks, we insert \textbf{prune decoder} between different ViT layers. \Cref{fig:VLP_arch}(a) presents the ViT model with prune decoder \textbf{F}$_{prune}$. Here we divide the segmentation model's ViT into $M$ pruning stages and put the prune decoder at the end of each stage. Each stage will have $l_{m}$ ViT layers. \textbf{F}$_{prune}$ takes the image tokens $T_{img}^{l_{m}}$ from the output of the $l_{m}$ ViT layers and  the $\langle$\text{SEG}$\rangle$ token $T_{seg}$ from the output of MLLM.
\begin{equation}
    P_m = \mathbf{F}_{prune}(T_{img}^{l_m}, T_{seg})
\end{equation}
The $P_{mn}$ ($n\in \{1,2,..,N\}$) is the relative score between $n^{\text{th}}$ image token $T_{img}^{l_m}[n]$ to the reasoning task. As shown in~\cref{fig:VLP_arch}(a), after generating the relative score of each image token, we pass the image tokens with top $ (1- r_{m}) \times N$ relative score to the next ViT layer while freezing the lowest $r_{m}\times N$ tokens. Here ``freeze'' means the rest $r_{m} \times N$ do not participate in the self-attention (\cref{eq:self-attention}) until the next pruning stage. Therefore, after pruning, only $1-r_{m}$ image tokens participate in the computation which significantly reduces the computation of ViT. In~\cref{sec:reactivation}, we discuss how we reuse those frozen tokens to improve the accuracy. 

\subsection{Prune Decoder Design and Training} \label{sec:prune_decoder}
~\Cref{fig:VLP_arch}(b) presents the model architecture of $\textbf{F}_{prune}$. Considering that the embedding dimension of image tokens (1280 for ViT-H) is generally higher than the $\langle$\text{SEG}$\rangle$ embedding dimension (256 for LLaVa), we first use a neck convolution layer to reduce the image token embedding from $D$ to $d$, where $D$ is the ViT image token embedding dimension and $d$ is the $\langle$\text{SEG}$\rangle$ embedding dimension. We then concatenate the $\langle$\text{SEG}$\rangle$ embedding vector $T_{seg}$ with a trainable query token embedding $T_{q}$ and perform self-attention of the concatenated vector.
\begin{equation}
    T_{cat} = FFN(MHSA(T_q | T_{seg}))
\end{equation}
The shape of $T_{cat}$ is $2 \times d$, where $T_{cat}[0]$ is the query token embedding and $T_{cat}[1]$ is $\langle$\text{SEG}$\rangle$ embedding. We then perform two-way cross attention between $T_{cat}$ and $T_{img}$.  
\begin{equation}
    T_{cat}' = FFN(MHSA(tgt=T_{cat}, src=T_{img}))
\end{equation}
\begin{equation}
    T_{img}' = MHSA(tgt=T_{img}, src=T_{cat}')
\end{equation}
The self-attention of $T_{q}|T_{seg}$ and cross attention between $T_{cat}$ and $T_{img}$ make the query token understand the relation between the image patch tokens and reasoning task. In the end, we perform inner-product between query embedding ($T_{q} = T_{cat}'[0]$) and image tokens ($T_{img}'$) to get the final image patch relation score:
\begin{equation}
    P_{s} = T_{img}' \times T_{cat}^{'\text{T}}[0]
\end{equation}
Suppose the pruning rate is $r_m$, we get the pruning mask $P_{mask}$ based on each image patch tokens' score
\begin{equation} \label{eq:prune}
P_{mask}[i] = 
\begin{cases} 
1 & \text{if } P_{s}[i] \geq \text{Top}_{r_m(P_{s})} \\
0 & \text{otherwise}
\end{cases}
\end{equation}
Here we use the Top$_{r_m(P_{s})}$ as the threshold value, which is the $(r_{m} \times N)^{\text{th}}$ highest score in $P$. As a result, roughly $(1-r_m)\times N$ of the image tokens will be retained. During the training stage, we use the Sigmoid function~\cite{han1995influence} to smooth out~\cref{eq:prune}. 
To train the prune decoder, we freeze both MLLM and ViT and only train the prune decoder and mask decoder with the training loss $L$:
\begin{equation}
\begin{split}
    L = & \, \underbrace{CE(X_{res}, X_{gt}) + DICE(X_{res}, X_{gt})}_{L_{X}} \\ & + \underbrace{CE(P_{mask}, P_{gt}) +DICE(P_{mask}, P_{gt})}_{L_{P}}
\end{split}
\end{equation}
Here $CE(\cdot, \cdot)$ and $DICE(\cdot, \cdot)$ represent the cross entropy loss and dice loss functions~\cite{jadon2020survey} respectively. The training loss includes two parts: final reasoning segmentation mask loss $L_{X}$ and pruning mask loss $L_{P}$. For the $L_{X}$, we follow previous work~\cite{kirillov2023segment}. We set the ground truth of pruning masks as:

\begin{flalign}
& P_{gt}\left(\text{idx}\right) = 
\begin{cases} 
1 & \text{if } \max(X_{gt}^{patch}(i,j)) > 0 \\
0 & \text{otherwise}
\end{cases}, & \\
& \text{where } \text{idx} = i \times \left(\frac{W}{P}\right) + j \quad \text{and} \quad \text{idx} \in [1,N] &
\end{flalign}

Here the $(i, j)$ is the patch index of the original ground truth mask $X_{gt}$. The ground truth pruning mask label $P_{gt}$ at a specific index is assigned a value of $1$ if the corresponding patch within the final mask includes at least one positive pixel; if not, it is assigned a value of $0$.

\subsection{Pruned Tokens Reactivation} \label{sec:reactivation}
As illustrated in~\cref{fig:VLP_arch}, we reactivate the pruned image tokens in two scenarios. Firstly, the pruned tokens from stage $m$ will be reactivated at stage $m+1$. This process aids the prune decoder in re-selecting image tokens that are most pertinent to reasoning tasks, thereby correcting any misdrops from earlier stages. Secondly, following the completion of all ViT layers, we will merge the pruned image tokens with the remaining tokens pertinent to reasoning tasks, and feed the combined tokens (with shape $N \times D$) into the mask decoder. This approach enhances the mask decoder's ability to reconstruct the final segmentation mask. 

\section{Experiments}
\subsection{Dataset and Metrics}
To verify the extensive applicability of VLTP for TOS tasks, we perform experiments on two datasets: RIO~\cite{qu2024rio} and COCO-Tasks~\cite{sawatzky2019object}. Both of these datasets are derived from MS COCO 2014~\cite{lin2014microsoft}. Specifically, RIO encompasses over 100 different tasks, and the entire dataset is categorized into two groups based on difficulty: \textit{common} and \textit{uncommon}. The COCO-Tasks dataset comprises 14 distinct tasks. Following the common convention, we utilize the mean intersection over union (mIoU) to assess the segmentation accuracy and the number of floating-point operations in giga (GFLOPs) to measure the model computational intensity. We perform all experiments using three NVIDIA RTX A6000 GPUs.

\subsection{Vision Language Fintuning}

\begin{table}[t]
\centering
\caption{Comparison with previous works on RIO~\cite{qu2024rio} dataset.}
\label{tab:accuracy_compare}
\resizebox{0.4\textwidth}{!}{
\begin{tabular}{lcc}
\toprule
\multirow{2}{*}{Model} & Common        & Uncommon      \\
                       & mIoU (\%)     & mIoU (\%)     \\ \midrule \midrule
MDETR~\cite{kamath2021mdetr}                  & 44.14         & 22.03         \\
TOIST~\cite{li2022toist}                  & 45.07         & 19.41         \\
Polyformer~\cite{liu2023polyformer}             & 48.75         & 26.77         \\
GROUNDHOG~\cite{zhang2024groundhog}              & 57.9          & 33.9          \\
SAM ViT-H              & \textbf{60.4} & \textbf{37.2} \\
SAM ViT-L              & 55.7          & 32.4          \\
SAM ViT-B              & 50.9          & 27.9          \\ \bottomrule
\end{tabular}%
}
\end{table}

\begin{table}[t]
\centering
\caption{Comparison of training scheme. All results are reported based on RIO common dataset. \ding{169} represents the extra finetune epochs.}
\label{tab:compare_training}
\resizebox{0.45\textwidth}{!}{
\begin{tabular}{lcc}
\toprule
Method                              & GFLOPs & mIoU (\%) \\ \midrule \midrule
Baseline                            & 2976   & 60.4      \\
+VLTP@Direct (\ding{169}0)                     & 2227   & 39.5      \\
+VLTP@Finetune ($F_{prune}$) (\ding{169}5)     & 2227   & 50.3      \\
+VLTP@Finetune ($F_{dec}$ + $F_{prune}$) (\ding{169}9) & 2227   & \textbf{60.1}      \\ \bottomrule
\end{tabular}%
}
\vspace{-3mm}
\end{table}

We selected LLaVA 7B~\cite{liu2024visual} to serve as MLLM. LLaVA processes both textual tasks and images as inputs to produce reasoning guidance $\langle$\text{SEG}$\rangle$. For the segmentation model, we opted for SAM, a well-known foundation model that enables prompt-guided segmentation. Following previous research~\cite{lai2024lisa, yan2024visa, qian2024affordancellm}, we perform end-to-end visual instruction fine-tuning on the entire system. For the LLaVA module, we utilize LoRA~\cite{hu2021lora} for efficient fine-tuning, while for the SAM module, we freeze the ViT backbone and only train the mask decoder.~\Cref{tab:accuracy_compare} shows the reasoning segmentation accuracy of the entire system on the RIO dataset, along with a comparison to earlier studies. Compared to the SOTA method~\cite{zhang2024groundhog}, guided by MLLM, the SAM with the ViT-H backbone model, exceeds it by 2.5\% on RIO common parts and 3.3\% on RIO uncommon parts. This result highlights the importance of integrating MLLM with the segmentation model in visual reasoning tasks.

\subsection{VLTP Framework Setup}

\begin{table*}[t]
\centering
\caption{Pruning ratio exploration for SAM ViT-H. Here we assume the pruning position is at layer 16 and layer 24 and both layer's pruning ratio is the same. All results are reported based on RIO common dataset.}
\label{tab:ratio_explore}
\resizebox{0.75\textwidth}{!}{
\begin{tabular}{cl|ccccccc}
\toprule
\multicolumn{2}{c|}{Pruning Rate (\%)} & 50   & 70   & 70   & 80   & 80    & 90    & 90   \\
\multicolumn{2}{c|}{Training scheme} & - & zero shot & finetune & zero shot & finetune & zero shot & finetune \\ \midrule \midrule
\multicolumn{2}{c|}{GFLOPs}            & 2227 & 1930 & \textbf{1930} & 1782 & \textbf{1782}  & 1636  & 1636 \\
\multicolumn{2}{c|}{Common (\%)}            & 60.1 & 57.1 & \textbf{59.6} & 52.1 & \textbf{59.4}  & 39.76 & 51.3 \\
\multicolumn{2}{c|}{Uncommon (\%)}          & 37.5 & 35   & \textbf{37.9} & 32.2 & \textbf{37.6} & 19.01 & 28.4 \\ \bottomrule
\end{tabular}%
}
\end{table*}

\begin{figure}[t]
  \centering \includegraphics[width=0.4\textwidth]{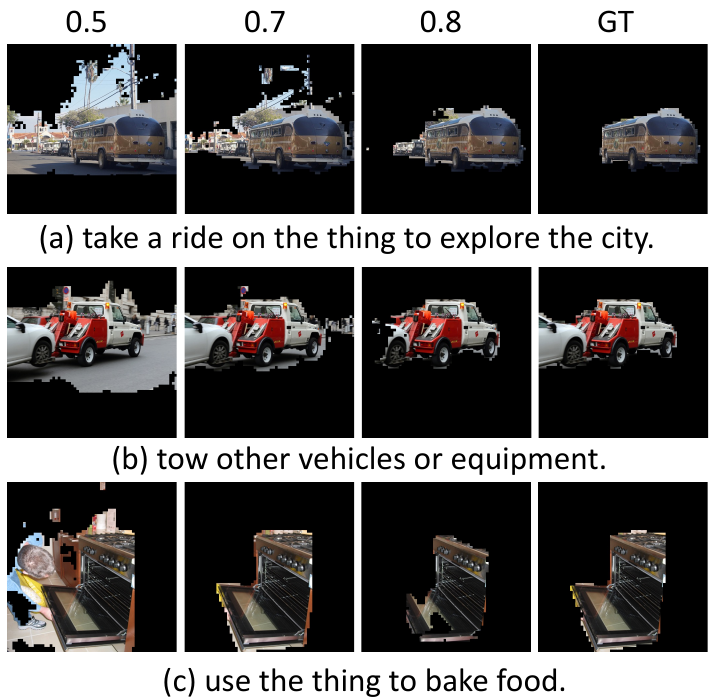}
   \caption{Visualization of VLTP image patch pruning for SAM ViT-H at layers 16 and 24. Three distinct pruning rates (0.5, 0.7, and 0.8) are illustrated alongside the ground truth (GT) task-related image patches.}
   \label{fig:visualize_prune_rate}
   \vspace{-2.em}
\end{figure}

To enhance the computational efficiency of SAM ViT, we integrate the prune decoder at different ViT layers. For SAM ViT-H, there are a total of 32 layers, with every 8 layers containing 7 layers of window attention and 1 layer of global attention. To more effectively model the relational importance between image tokens and reasoning $\langle$\text{SEG}$\rangle$, we insert the prune decoder after the global attention layer. Consequently, for SAM ViT-H, we place the prune decoder at layers 8, 16, and 24. It's important to note that the prune decoder only involves query embedding self-attention and two-way query embedding-to-image patch embedding cross-attention. \textbf{Thus, its computation is merely 6 GFLOPs, significantly less than the computation of ViT-H, which exceeds 2900 GFLOPs.} Another crucial point is that there is only a single prune decoder model within the entire VLTP framework. However, due to the reactivation mechanism, the same prune decoder may be called at multiple layers.

\begin{table}[t]
\centering
\caption{Investigation of SAM ViT-H pruning locations. The table presents the maximum accuracy and the most efficient outcome (measured in GFLOPs) for each pruning position combination. All outcomes are derived from the RIO common dataset. Each invocation of the prune decoder adds an additional 6 GFLOPs.}
\label{tab:position_explore}
\resizebox{0.81\columnwidth}{!}{
\begin{tabular}{cccc}
\toprule
Position      & Pruning Rate   & GFLOPS & mIoU (\%)   \\ \midrule \midrule
Baseline      & \{0\}          & 2976   & 60.4        \\ \midrule
\{8\}         & \{20\}         & 2529   & 59.8 (-0.6) \\
\{8\}         & \{40\}         & 2083   & 53.1 (-7.3) \\ \midrule
\{8, 16\}     & \{20, 40\}     & 2232   & 58.8 (-1.6) \\
\{8, 16\}     & \{20, 60\}     & 1783   & 53.7 (-6.7) \\ \midrule
\{16, 24\}    & \{50, 50\}     & 2227   & 60.1 (-0.3) \\
\{16, 24\}    & \{80, 80\}     & 1782   & 58.9 (-1.2) \\ \midrule
\{8, 16, 24\} & \{20, 40, 40\} & 2232   & 58.9 (-1.5) \\
\{8, 16, 24\} & \{20, 40, 60\} & 1853   & 53 (-7.4)   \\
\{8, 16, 24\} & \{50, 50, 50\} & 1860   & 53.3 (-7.1) \\ \midrule
\end{tabular}}
\end{table}

\Cref{tab:compare_training} demonstrates the impact of different training strategies on final pruning outcomes. We assume a pruning rate of 50\%, inserting the prune decoder at layers 16 and 24. This implies that starting from layer 16, only 50\% of image tokens will be involved in the SAM ViT computation. The results indicate that when both $F_{prune}$ and $F_{dec}$ are fine-tuned simultaneously, VLTP reduces the SAM ViT's computation from 2976 GFLOPs to 2227 GFLOPs, yielding approximately a 25\% reduction in GFLOPs with only a 0.3\% drop in mIoU. Consequently, in the subsequent sections, both the prune decoder and mask decoder will be fine-tuned to achieve optimal pruning results.

\begin{table}[t]
\centering
\caption{Ablation study for effect's of pruned tokens reevaluation. Here the model is SAM ViT-H and dataset is RIO common. Each invo-
cation of the prune decoder adds an additional 6 GFLOPs.}
\label{tab:ablation_reactivation}
\resizebox{0.81\columnwidth}{!}{%
\begin{tabular}{cccc}
\toprule
Position   & Pruning Rate & GFLOPS & mIoU (\%)   \\ \midrule \midrule
\{16\}     & \{50\}      & 2221   & 59.8        \\
\{16, 24\} & \{50, 50\} & 2227   & 60.1 (+0.3) \\ \midrule
\{16\}     & \{70\}      & 1923   & 58.2        \\
\{16, 24\} & \{70, 70\} & 1930   & 59.6 (+1.4) \\ \midrule
\{16\}     & \{80\}      & 1775   & 56.8        \\
\{16, 24\} & \{80, 80\} & 1782   & 59.4 (+2.6) \\ \bottomrule
\end{tabular}}
\end{table}

\begin{table}[t]
\centering
\caption{A comparison of various pruning techniques. Each method is applied on SAM ViT-H. Random 50\% indicates randomly dropping 50\% of image tokens starting from layer 16.}
\label{tab:method_compare}
\resizebox{0.81\columnwidth}{!}{%
\begin{tabular}{cccc}
\toprule
Method                    & VLM support  & GFLOPs        & mIoU                 \\ \midrule \midrule
Baseline                  & No           & 2976          & 60.4 (-0)                 \\
Random 50\%               & No           & 2297          & 38.2 (-22.2)         \\
CTS~\cite{lu2023content}                       & No           & 2232          & 35.7 (-24.7)         \\
DToP~\cite{tang2023dynamic}                      & No           & 1892          & 36.3 (-24.1)         \\
SViT~\cite{liu2024revisiting}                      & No           & 1935          & 33.5 (-26.9)         \\ \midrule
\textbf{VLTP (this work)} & \textbf{Yes} & \textbf{1782} & \textbf{59.4 (-1.0)} \\ \bottomrule
\end{tabular}%
}
\vspace{-3mm}
\end{table}

\begin{table*}[t]
\centering
\caption{Main results on two different TOS benchmarks.}
\label{tab:main_res}
\resizebox{0.8\textwidth}{!}{%
\begin{tabular}{lccccccc}
\toprule
\multirow{2}{*}{Model} &
  \multicolumn{1}{l}{\multirow{2}{*}{pruning rate}} &
  \multicolumn{2}{c}{RIO common} &
  \multicolumn{2}{c}{RIO uncommon} &
  \multicolumn{2}{c}{COCO-Tasks} \\
              & \multicolumn{1}{l}{} & mIoU(\%) & GFLOPs & mIoU(\%) & GFLOPs & mIoU(\%) & GFLOPs \\ \midrule \midrule
SAM ViT-H     & -                    & 60.4     & 2976   & 37.2     & 2976   & 44.6     & 2976   \\
VLTP@Finetune & 0.5                  & 60.1     & 2227   & 37.5     & 2224   & 44.2     & 2219   \\
VLTP@Finetune & 0.8                  & 59.4     & 1782   & 37.6    & 1774   & 43.2     & 1771   \\ \midrule
SAM ViT-L     & -                    & 55.7     & 1491   & 32.4     & 1491   & 40.1     & 1491   \\
VLTP@Finetune & 0.5                  & 55.1     & 1118   & 32.1     & 1121   & 39.6     & 1127   \\
VLTP@Finetune & 0.8                  & 54.3     & 895    & 31.4     & 901    & 38.7     & 905    \\ \bottomrule
\end{tabular}%
}
\end{table*}
 
\subsection{Exploration of Pruning Rate and Position}

\begin{figure}[t]
  \centering \includegraphics[width=0.4\textwidth]{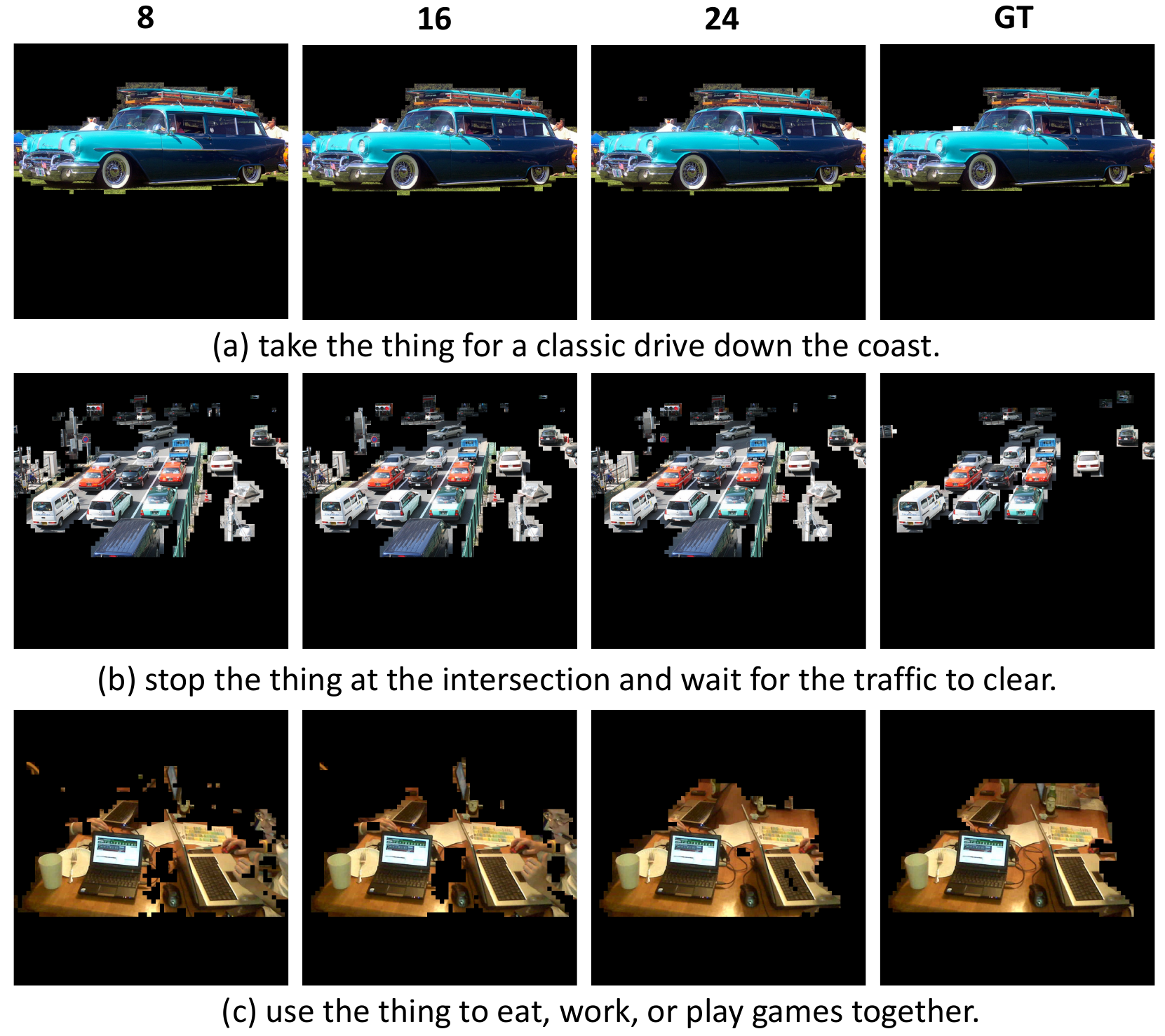}
   \caption{Visualization of VLTP image patch pruning for SAM ViT-H at layers 8, 16, and 24 along with the ground truth (GT). The pruning rate is 0.7.}
   \label{fig:visualize_position}
   \vspace{-5mm}
\end{figure}

\begin{figure*}[t]
  \centering \includegraphics[width=0.9\textwidth]{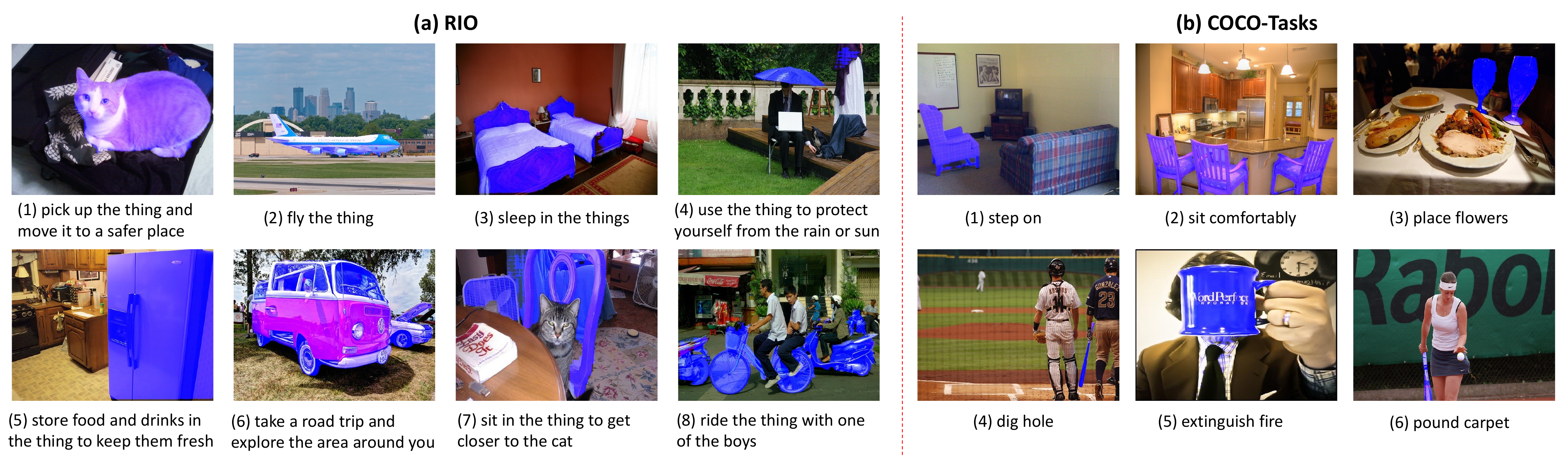}
   \caption{Visualized results. The segmentation results are predicted on RIO (left) and COCO-Tasks (right). The model is SAM ViT-H under the guidance of LLaVA 7B with VLTP@Finetune. We set the pruning position at layers 16 and 24 with the pruning rate as 50\%.}
   \label{fig:visualize_final}
\end{figure*}

\Cref{tab:ratio_explore} illustrates the effects of pruning as the pruning rate $r_m$ increases. In this study, we set the pruning positions of SAM ViT at layers 16 and 24. Instead of finetuning the pruning and mask decoders from scratch, we base our experiments on a pre-trained model with a 50\% pruning rate, as depicted in~\cref{tab:compare_training}. Initially, we increase the pruning rate without any finetuning, and the results in~\cref{tab:ratio_explore} indicate an accuracy drop. Consequently, we subsequently finetune both the prune decoder and mask decoder together after adjusting the pruning rate. The experiment demonstrates that VLTP achieves approximately 35\% GFLOPs with just a 0.8\% mIoU reduction at a 70\% pruning rate, and around 40\% GFLOPs with only a 1\% mIoU reduction at an 80\% pruning rate, compared to the original SAM ViT-H. Additionally, we display the image patch-dropping visual results in ~\cref{fig:visualize_prune_rate}. Under MLLM guidance, VLTP retains the most essential image patches while freezing irrelevant ones, in comparison to the ground truth.  

In~\cref{tab:position_explore}, we illustrate the influence of different pruning positions on the ultimate pruning accuracy. The findings suggest that initiating pruning from layer 8 leads to a greater accuracy reduction compared to starting from layer 16, assuming the same level of ViT computation reduction. We attribute this to inadequate image feature extraction. Image patches that aren't directly pertinent to reasoning tasks still contribute to the final mask generation. Consequently, removing these patches at an earlier layer can negatively impact the final mask generation. ~\Cref{fig:visualize_position} visualizes the patch dropping at various layer positions. 

\subsection{Ablation Study and Final Results}

\Cref{tab:ablation_reactivation} illustrates the impact of reevaluating pruned image tokens. We compare the final reasoning segmentation accuracy between pruning only at layer 16 and pruning at both layer 16 and layer 24. Reassessing token relevance at layer 24 enhances the accuracy of the final reasoning. This is due to the prune decoder potentially making incorrect predictions about the relevance of image tokens to the reasoning task at earlier stages. Hence, it is crucial to reassess the importance of these tokens at deeper ViT layers. In~\cref{tab:method_compare}, we compare various pruning methods for task-oriented reasoning tasks. Although prior research~\cite{lu2023content, tang2023dynamic, liu2024revisiting} has shown effectiveness in pruning ViTs for traditional semantic or instance segmentation tasks, these methods are not suitable for TOS as they lack vision-language guidance. For instance, the policy network in CTS~\cite{lu2023content} is unable to accurately predict whether four image tokens belong to the same semantic class because each token's class can change depending on the reasoning task. Consequently, the pruning networks proposed in DToP~\cite{tang2023dynamic} and SViT~\cite{liu2024revisiting} are also ineffective as they do not incorporate reasoning guidance. VLTP addresses these limitations by integrating both ViT image tokens and MLLM reasoning guidance, thus achieving superior performance. In~\cref{tab:main_res}, we provide a summary of the results when applying VLTP to SAM ViT-H and SAM ViT-L for the RIO common, RIO uncommon, and COCO-Tasks datasets. VLTP effectively maintains task-relevant image tokens while significantly reducing SAM ViT's computational costs. On average, VLTP reduces SAM ViT's computation by 40\% with only a 1\% loss in mIoU. \Cref{fig:visualize_final} showcases our framework's segmentation visualized results on the RIO and COCO-Task datasets.

\section{Conclusion}
In this work, we present a technique for vision language-guided token pruning in ViTs dedicated to TOS. Recognizing that not every image token needs to be processed by all layers of the ViT, we introduce a novel prune decoder to estimate the relevance of each image patch token to the reasoning task. After evaluation, we retain those image tokens with high relevance to the task and skip the rest. To mitigate incorrect predictions in the initial ViT layers, we also propose a reactivation mechanism that reassesses the relevance of all image tokens to the task. Comprehensive experimental results indicate that our method enhances model efficiency by significantly reducing the computational load of ViTs while producing state-of-the-art segmentation results.

\section*{Acknowledgements}
This work was supported in part by the DARPA Young Faculty Award, the National Science Foundation (NSF) under Grants \#2127780, \#2319198, \#2321840, \#2312517, and \#2235472, the Semiconductor Research Corporation (SRC), the Office of Naval Research through the Young Investigator Program Award, and Grants \#N00014-21-1-2225 and N00014-22-1-2067. Additionally, support was provided by the Air Force Office of Scientific Research under Award \#FA9550-22-1-0253, along with generous gifts from Xilinx and Cisco.

%%%%%%%%% REFERENCES
{\small
\bibliographystyle{ieee_fullname}
\bibliography{egbib}

\begin{thebibliography}{10}\itemsep=-1pt

\bibitem{ba2016layer}
JL Ba.
\newblock Layer normalization.
\newblock {\em arXiv preprint arXiv:1607.06450}, 2016.

\bibitem{bolya2019yolact}
Daniel Bolya, Chong Zhou, Fanyi Xiao, and Yong~Jae Lee.
\newblock Yolact: Real-time instance segmentation.
\newblock In {\em Proceedings of the IEEE/CVF international conference on computer vision}, pages 9157--9166, 2019.

\bibitem{caesar2018coco}
Holger Caesar, Jasper Uijlings, and Vittorio Ferrari.
\newblock Coco-stuff: Thing and stuff classes in context.
\newblock In {\em Proceedings of the IEEE conference on computer vision and pattern recognition}, pages 1209--1218, 2018.

\bibitem{caron2021emerging}
Mathilde Caron, Hugo Touvron, Ishan Misra, Herv{\'e} J{\'e}gou, Julien Mairal, Piotr Bojanowski, and Armand Joulin.
\newblock Emerging properties in self-supervised vision transformers.
\newblock In {\em Proceedings of the IEEE/CVF international conference on computer vision}, pages 9650--9660, 2021.

\bibitem{chen2024comprehensive}
Feiyang Chen, Ziqian Luo, Lisang Zhou, Xueting Pan, and Ying Jiang.
\newblock Comprehensive survey of model compression and speed up for vision transformers.
\newblock {\em arXiv preprint arXiv:2404.10407}, 2024.

\bibitem{chen2024taskclip}
Hanning Chen, Wenjun Huang, Yang Ni, Sanggeon Yun, Fei Wen, Hugo Latapie, and Mohsen Imani.
\newblock Taskclip: Extend large vision-language model for task oriented object detection.
\newblock {\em arXiv preprint arXiv:2403.08108}, 2024.

\bibitem{cheng2021per}
Bowen Cheng, Alex Schwing, and Alexander Kirillov.
\newblock Per-pixel classification is not all you need for semantic segmentation.
\newblock {\em Advances in neural information processing systems}, 34:17864--17875, 2021.

\bibitem{cuttano2024does}
Claudia Cuttano, Gabriele Rosi, Gabriele Trivigno, and Giuseppe Averta.
\newblock What does clip know about peeling a banana?
\newblock In {\em Proceedings of the IEEE/CVF Conference on Computer Vision and Pattern Recognition}, pages 2238--2247, 2024.

\bibitem{dosovitskiy2020image}
Alexey DOSOVITSKIY.
\newblock An image is worth 16x16 words: Transformers for image recognition at scale.
\newblock {\em arXiv preprint arXiv:2010.11929}, 2020.

\bibitem{han1995influence}
Jun Han and Claudio Moraga.
\newblock The influence of the sigmoid function parameters on the speed of backpropagation learning.
\newblock In {\em International workshop on artificial neural networks}, pages 195--201. Springer, 1995.

\bibitem{he2023fastinst}
Junjie He, Pengyu Li, Yifeng Geng, and Xuansong Xie.
\newblock Fastinst: A simple query-based model for real-time instance segmentation.
\newblock In {\em Proceedings of the IEEE/CVF conference on computer vision and pattern recognition}, pages 23663--23672, 2023.

\bibitem{he2022masked}
Kaiming He, Xinlei Chen, Saining Xie, Yanghao Li, Piotr Doll{\'a}r, and Ross Girshick.
\newblock Masked autoencoders are scalable vision learners.
\newblock In {\em Proceedings of the IEEE/CVF conference on computer vision and pattern recognition}, pages 16000--16009, 2022.

\bibitem{he2016deep}
Kaiming He, Xiangyu Zhang, Shaoqing Ren, and Jian Sun.
\newblock Deep residual learning for image recognition.
\newblock In {\em Proceedings of the IEEE conference on computer vision and pattern recognition}, pages 770--778, 2016.

\bibitem{hu2021lora}
Edward~J Hu, Yelong Shen, Phillip Wallis, Zeyuan Allen-Zhu, Yuanzhi Li, Shean Wang, Lu Wang, and Weizhu Chen.
\newblock Lora: Low-rank adaptation of large language models.
\newblock {\em arXiv preprint arXiv:2106.09685}, 2021.

\bibitem{huang2024recoverable}
Wenjun Huang, Yang Ni, Arghavan Rezvani, SungHeon Jeong, Hanning Chen, Yezi Liu, Fei Wen, and Mohsen Imani.
\newblock Recoverable anonymization for pose estimation: A privacy-enhancing approach.
\newblock {\em arXiv preprint arXiv:2409.02715}, 2024.

\bibitem{huang2024intelligent}
Wenjun Huang, Arghavan Rezvani, Hanning Chen, Yang Ni, Sanggeon Yun, Sungheon Jeong, Guangyi Zhang, and Mohsen Imani.
\newblock Intelligent sensing framework: Near-sensor machine learning for efficient data transmission.
\newblock {\em IEEE Sensors Journal}, 2024.

\bibitem{jadon2020survey}
Shruti Jadon.
\newblock A survey of loss functions for semantic segmentation.
\newblock In {\em 2020 IEEE conference on computational intelligence in bioinformatics and computational biology (CIBCB)}, pages 1--7. IEEE, 2020.

\bibitem{jiang2021semantic}
Du Jiang, Gongfa Li, Chong Tan, Li Huang, Ying Sun, and Jianyi Kong.
\newblock Semantic segmentation for multiscale target based on object recognition using the improved faster-rcnn model.
\newblock {\em Future Generation Computer Systems}, 123:94--104, 2021.

\bibitem{kamath2021mdetr}
Aishwarya Kamath, Mannat Singh, Yann LeCun, Gabriel Synnaeve, Ishan Misra, and Nicolas Carion.
\newblock Mdetr-modulated detection for end-to-end multi-modal understanding.
\newblock In {\em Proceedings of the IEEE/CVF international conference on computer vision}, pages 1780--1790, 2021.

\bibitem{katharopoulos2020transformers}
Angelos Katharopoulos, Apoorv Vyas, Nikolaos Pappas, and Fran{\c{c}}ois Fleuret.
\newblock Transformers are rnns: Fast autoregressive transformers with linear attention.
\newblock In {\em International conference on machine learning}, pages 5156--5165. PMLR, 2020.

\bibitem{khan2022transformers}
Salman Khan, Muzammal Naseer, Munawar Hayat, Syed~Waqas Zamir, Fahad~Shahbaz Khan, and Mubarak Shah.
\newblock Transformers in vision: A survey.
\newblock {\em ACM computing surveys (CSUR)}, 54(10s):1--41, 2022.

\bibitem{kirillov2023segment}
Alexander Kirillov, Eric Mintun, Nikhila Ravi, Hanzi Mao, Chloe Rolland, Laura Gustafson, Tete Xiao, Spencer Whitehead, Alexander~C Berg, Wan-Yen Lo, et~al.
\newblock Segment anything.
\newblock In {\em Proceedings of the IEEE/CVF International Conference on Computer Vision}, pages 4015--4026, 2023.

\bibitem{kong2022spvit}
Zhenglun Kong, Peiyan Dong, Xiaolong Ma, Xin Meng, Wei Niu, Mengshu Sun, Xuan Shen, Geng Yuan, Bin Ren, Hao Tang, et~al.
\newblock Spvit: Enabling faster vision transformers via latency-aware soft token pruning.
\newblock In {\em European conference on computer vision}, pages 620--640. Springer, 2022.

\bibitem{lai2024lisa}
Xin Lai, Zhuotao Tian, Yukang Chen, Yanwei Li, Yuhui Yuan, Shu Liu, and Jiaya Jia.
\newblock Lisa: Reasoning segmentation via large language model.
\newblock In {\em Proceedings of the IEEE/CVF Conference on Computer Vision and Pattern Recognition}, pages 9579--9589, 2024.

\bibitem{li2022toist}
Pengfei Li, Beiwen Tian, Yongliang Shi, Xiaoxue Chen, Hao Zhao, Guyue Zhou, and Ya-Qin Zhang.
\newblock Toist: Task oriented instance segmentation transformer with noun-pronoun distillation.
\newblock {\em Advances in Neural Information Processing Systems}, 35:17597--17611, 2022.

\bibitem{li2023vit}
Zhikai Li and Qingyi Gu.
\newblock I-vit: Integer-only quantization for efficient vision transformer inference.
\newblock In {\em Proceedings of the IEEE/CVF International Conference on Computer Vision}, pages 17065--17075, 2023.

\bibitem{lin2014microsoft}
Tsung-Yi Lin, Michael Maire, Serge Belongie, James Hays, Pietro Perona, Deva Ramanan, Piotr Doll{\'a}r, and C~Lawrence Zitnick.
\newblock Microsoft coco: Common objects in context.
\newblock In {\em Computer Vision--ECCV 2014: 13th European Conference, Zurich, Switzerland, September 6-12, 2014, Proceedings, Part V 13}, pages 740--755. Springer, 2014.

\bibitem{liu2024visual}
Haotian Liu, Chunyuan Li, Qingyang Wu, and Yong~Jae Lee.
\newblock Visual instruction tuning.
\newblock {\em Advances in neural information processing systems}, 36, 2024.

\bibitem{liu2023polyformer}
Jiang Liu, Hui Ding, Zhaowei Cai, Yuting Zhang, Ravi~Kumar Satzoda, Vijay Mahadevan, and R Manmatha.
\newblock Polyformer: Referring image segmentation as sequential polygon generation.
\newblock In {\em Proceedings of the IEEE/CVF Conference on Computer Vision and Pattern Recognition}, pages 18653--18663, 2023.

\bibitem{liu2024revisiting}
Yifei Liu, Mathias Gehrig, Nico Messikommer, Marco Cannici, and Davide Scaramuzza.
\newblock Revisiting token pruning for object detection and instance segmentation.
\newblock In {\em Proceedings of the IEEE/CVF Winter Conference on Applications of Computer Vision}, pages 2658--2668, 2024.

\bibitem{lu2023content}
Chenyang Lu, Daan de Geus, and Gijs Dubbelman.
\newblock Content-aware token sharing for efficient semantic segmentation with vision transformers.
\newblock In {\em Proceedings of the IEEE/CVF Conference on Computer Vision and Pattern Recognition}, pages 23631--23640, 2023.

\bibitem{lu202012}
Jiasen Lu, Vedanuj Goswami, Marcus Rohrbach, Devi Parikh, and Stefan Lee.
\newblock 12-in-1: Multi-task vision and language representation learning.
\newblock In {\em Proceedings of the IEEE/CVF conference on computer vision and pattern recognition}, pages 10437--10446, 2020.

\bibitem{ma2024groma}
Chuofan Ma, Yi Jiang, Jiannan Wu, Zehuan Yuan, and Xiaojuan Qi.
\newblock Groma: Localized visual tokenization for grounding multimodal large language models.
\newblock {\em arXiv preprint arXiv:2404.13013}, 2024.

\bibitem{mahmud2024papr}
Tanvir Mahmud, Burhaneddin Yaman, Chun-Hao Liu, and Diana Marculescu.
\newblock Papr: Training-free one-step patch pruning with lightweight convnets for faster inference.
\newblock {\em arXiv preprint arXiv:2403.16020}, 2024.

\bibitem{mottaghi2014role}
Roozbeh Mottaghi, Xianjie Chen, Xiaobai Liu, Nam-Gyu Cho, Seong-Whan Lee, Sanja Fidler, Raquel Urtasun, and Alan Yuille.
\newblock The role of context for object detection and semantic segmentation in the wild.
\newblock In {\em Proceedings of the IEEE conference on computer vision and pattern recognition}, pages 891--898, 2014.

\bibitem{pan2021ia}
Bowen Pan, Rameswar Panda, Yifan Jiang, Zhangyang Wang, Rogerio Feris, and Aude Oliva.
\newblock Ia-red$^2$: Interpretability-aware redundancy reduction for vision transformers.
\newblock {\em Advances in Neural Information Processing Systems}, 34:24898--24911, 2021.

\bibitem{qian2024affordancellm}
Shengyi Qian, Weifeng Chen, Min Bai, Xiong Zhou, Zhuowen Tu, and Li~Erran Li.
\newblock Affordancellm: Grounding affordance from vision language models.
\newblock In {\em Proceedings of the IEEE/CVF Conference on Computer Vision and Pattern Recognition}, pages 7587--7597, 2024.

\bibitem{qu2024rio}
Mengxue Qu, Yu Wu, Wu Liu, Xiaodan Liang, Jingkuan Song, Yao Zhao, and Yunchao Wei.
\newblock Rio: A benchmark for reasoning intention-oriented objects in open environments.
\newblock {\em Advances in Neural Information Processing Systems}, 36, 2024.

\bibitem{rao2021dynamicvit}
Yongming Rao, Wenliang Zhao, Benlin Liu, Jiwen Lu, Jie Zhou, and Cho-Jui Hsieh.
\newblock Dynamicvit: Efficient vision transformers with dynamic token sparsification.
\newblock {\em Advances in neural information processing systems}, 34:13937--13949, 2021.

\bibitem{sawatzky2019object}
Johann Sawatzky, Yaser Souri, Christian Grund, and Jurgen Gall.
\newblock What object should i use?-task driven object detection.
\newblock In {\em Proceedings of the IEEE/CVF Conference on Computer Vision and Pattern Recognition}, pages 7605--7614, 2019.

\bibitem{shang2024llava}
Yuzhang Shang, Mu Cai, Bingxin Xu, Yong~Jae Lee, and Yan Yan.
\newblock Llava-prumerge: Adaptive token reduction for efficient large multimodal models.
\newblock {\em arXiv preprint arXiv:2403.15388}, 2024.

\bibitem{strudel2021segmenter}
Robin Strudel, Ricardo Garcia, Ivan Laptev, and Cordelia Schmid.
\newblock Segmenter: Transformer for semantic segmentation.
\newblock In {\em Proceedings of the IEEE/CVF international conference on computer vision}, pages 7262--7272, 2021.

\bibitem{tang2023cotdet}
Jiajin Tang, Ge Zheng, Jingyi Yu, and Sibei Yang.
\newblock Cotdet: Affordance knowledge prompting for task driven object detection.
\newblock In {\em Proceedings of the IEEE/CVF International Conference on Computer Vision}, pages 3068--3078, 2023.

\bibitem{tang2023dynamic}
Quan Tang, Bowen Zhang, Jiajun Liu, Fagui Liu, and Yifan Liu.
\newblock Dynamic token pruning in plain vision transformers for semantic segmentation.
\newblock In {\em Proceedings of the IEEE/CVF International Conference on Computer Vision}, pages 777--786, 2023.

\bibitem{vaswani2017attention}
A Vaswani.
\newblock Attention is all you need.
\newblock {\em Advances in Neural Information Processing Systems}, 2017.

\bibitem{wang2023cut}
Xudong Wang, Rohit Girdhar, Stella~X Yu, and Ishan Misra.
\newblock Cut and learn for unsupervised object detection and instance segmentation.
\newblock In {\em Proceedings of the IEEE/CVF conference on computer vision and pattern recognition}, pages 3124--3134, 2023.

\bibitem{wang2021end}
Yuqing Wang, Zhaoliang Xu, Xinlong Wang, Chunhua Shen, Baoshan Cheng, Hao Shen, and Huaxia Xia.
\newblock End-to-end video instance segmentation with transformers.
\newblock In {\em Proceedings of the IEEE/CVF conference on computer vision and pattern recognition}, pages 8741--8750, 2021.

\bibitem{xie2021segformer}
Enze Xie, Wenhai Wang, Zhiding Yu, Anima Anandkumar, Jose~M Alvarez, and Ping Luo.
\newblock Segformer: Simple and efficient design for semantic segmentation with transformers.
\newblock {\em Advances in neural information processing systems}, 34:12077--12090, 2021.

\bibitem{yan2024visa}
Cilin Yan, Haochen Wang, Shilin Yan, Xiaolong Jiang, Yao Hu, Guoliang Kang, Weidi Xie, and Efstratios Gavves.
\newblock Visa: Reasoning video object segmentation via large language models.
\newblock {\em arXiv preprint arXiv:2407.11325}, 2024.

\bibitem{yang2023global}
Huanrui Yang, Hongxu Yin, Maying Shen, Pavlo Molchanov, Hai Li, and Jan Kautz.
\newblock Global vision transformer pruning with hessian-aware saliency.
\newblock In {\em Proceedings of the IEEE/CVF conference on computer vision and pattern recognition}, pages 18547--18557, 2023.

\bibitem{ye2024voco}
Xubing Ye, Yukang Gan, Xiaoke Huang, Yixiao Ge, Ying Shan, and Yansong Tang.
\newblock Voco-llama: Towards vision compression with large language models.
\newblock {\em arXiv preprint arXiv:2406.12275}, 2024.

\bibitem{yin2022vit}
Hongxu Yin, Arash Vahdat, Jose~M Alvarez, Arun Mallya, Jan Kautz, and Pavlo Molchanov.
\newblock A-vit: Adaptive tokens for efficient vision transformer.
\newblock In {\em Proceedings of the IEEE/CVF conference on computer vision and pattern recognition}, pages 10809--10818, 2022.

\bibitem{yu2016modeling}
Licheng Yu, Patrick Poirson, Shan Yang, Alexander~C Berg, and Tamara~L Berg.
\newblock Modeling context in referring expressions.
\newblock In {\em Computer Vision--ECCV 2016: 14th European Conference, Amsterdam, The Netherlands, October 11-14, 2016, Proceedings, Part II 14}, pages 69--85. Springer, 2016.

\bibitem{zhang2024advancing}
Guangyi Zhang and Wenjun Huang.
\newblock Advancing circuit transient response macromodeling: From conventional neural networks to siamese-lstm.
\newblock {\em IEEE Transactions on Circuits and Systems I: Regular Papers}, 2024.

\bibitem{zhang2022glipv2}
Haotian Zhang, Pengchuan Zhang, Xiaowei Hu, Yen-Chun Chen, Liunian Li, Xiyang Dai, Lijuan Wang, Lu Yuan, Jenq-Neng Hwang, and Jianfeng Gao.
\newblock Glipv2: Unifying localization and vision-language understanding.
\newblock {\em Advances in Neural Information Processing Systems}, 35:36067--36080, 2022.

\bibitem{zhang2024groundhog}
Yichi Zhang, Ziqiao Ma, Xiaofeng Gao, Suhaila Shakiah, Qiaozi Gao, and Joyce Chai.
\newblock Groundhog: Grounding large language models to holistic segmentation.
\newblock In {\em Proceedings of the IEEE/CVF conference on computer vision and pattern recognition}, pages 14227--14238, 2024.

\bibitem{zhang2024efficientvit}
Zhuoyang Zhang, Han Cai, and Song Han.
\newblock Efficientvit-sam: Accelerated segment anything model without performance loss.
\newblock {\em arXiv preprint arXiv:2402.05008}, 2024.

\bibitem{zheng2021rethinking}
Sixiao Zheng, Jiachen Lu, Hengshuang Zhao, Xiatian Zhu, Zekun Luo, Yabiao Wang, Yanwei Fu, Jianfeng Feng, Tao Xiang, Philip~HS Torr, et~al.
\newblock Rethinking semantic segmentation from a sequence-to-sequence perspective with transformers.
\newblock In {\em Proceedings of the IEEE/CVF conference on computer vision and pattern recognition}, pages 6881--6890, 2021.

\bibitem{zhou2024tinyllava}
Baichuan Zhou, Ying Hu, Xi Weng, Junlong Jia, Jie Luo, Xien Liu, Ji Wu, and Lei Huang.
\newblock Tinyllava: A framework of small-scale large multimodal models.
\newblock {\em arXiv preprint arXiv:2402.14289}, 2024.

\end{thebibliography}
}

\end{document}